# Demand-Driven Asset Reutilization Analytics

Abbas Raza Ali
BAO Center of Competence, IBM
London, UK
abbas.raza.ali@gmail.com

Dr. Pitipong J. Lin
Smarter Supply Chain Analytics, IBM
Cambridge, MA, USA
pitipong@us.ibm.com

*Abstract*—Manufacturers have long benefited from reusing returned products and parts. This benevolent approach can minimize cost and help the manufacturer to play a role in sustaining the environment, something which is of utmost importance these days because of growing environment concerns. Reuse of returned parts and products aids environment sustainability because doing so helps reduce the use of raw materials, eliminate energy use to produce new parts, and minimize waste materials. However, handling returns effectively and efficiently can be difficult if the processes do not provide the visibility that is necessary to track, manage, and re-use the returns.

This paper applies advanced analytics on procurement data to increase reutilization in new build by optimizing Equal-to-New (ETN) parts return. This will reduce 'the spend' on new buy parts for building new product units. The process involves forecasting and matching returns supply to demand for new build. Complexity in the process is the forecasting and matching while making sure a reutilization engineering process is available. Also, this will identify high demand/value/yield parts for development engineering to focus.

Reutilization of returns is good for the environment. The higher the quantity of ETN and certified service parts (CSP) that are recovered and reused, the better the result is for the environment. So another major goal of this work is to minimize the solid waste volume that is disposed of from scrapped parts.

Analytics has been applied on different levels to enhance the optimization process including forecast of upgraded parts. Machine Learning algorithms are used to build an automated infrastructure that can support the transformation of ETN parts utilization in the procurement parts planning process. This system incorporate returns forecast in the planning cycle to reduce suppliers liability from 9 weeks to 12 months planning cycle, e.g., reduce 5% of 10 million US dollars liability.

I. INTRODUCTION

High-end electronics manufacturers have long benefited from reusing returned hardware products and parts. Reutilization of returns is important because it has a significant positive impact on the company's financial bottom line, e.g., high value parts such as processors and memory that are used in mainframe servers can yield up to a 70% margin when recovered and reused.

As part of continual improvement, an asset reutilization project was initiated to handle returned products and parts more efficiently. This project applied business analytics to various aspects of asset reutilization, such as forecasting the return of material and enabling optimized reutilization. It helps the manufacturers to better manage their returned goods, i.e., computer server machines for the case study discussed in this paper. They knew that these returns could be recycled into hardware products that are being manufactured or for service. By using Analytics, the behavior of the historic server market was analyzed to produce a forecast of returning products and parts. By using these forecasts, we can further analyze where and how to use those returns. The ability to reuse and recycle returns creates a demand-driven reverse supply chain which helps a company in finding ways to reduce its costs by reusing returned products and parts.

The ETN parts process takes back a part from customers and goes through a rigorous ETN path process of qualification, including lifespan and part usage analysis, testing, and reconditioning according to the engineering specifications before reuse [12].

The growing volume and range of information encountered in the reverse supply chain drove complexity and created a need for Business Analytics and Optimization (BAO). BAO capabilities are used to tackle problems in the original reverse supply chain. Following examples illustrate the above statement:

1) Descriptive analytics provides information about the past state or performance of the business and its environment.

2) Predictive analytics helps to predict (based on data and statistical techniques) with confidence what will happen next so that well-informed decisions can be made to improve overall business outcomes.

3) Prescriptive analytics recommends high-value alternative actions or decisions given a complex set of targets, limits and choices. Specially, optimization is used to examine how best outcome can be achieved for a particular situation.

The process of forecasting returns identifies the quantity of parts that will be returned in the next 12 months. Predictive Analytics provides the capability to find trends and extrapolate them in the future, e.g., this work helps to predict the future value of returns. The forecast enables collaboration with demand owners, such as Procurement team, who buy parts for manufacturing or service.

The ability to forecast returns, sufficiently in-advance, prevents procurement from incurring inventory liability when ordering new parts. Procurement purchases parts before they are needed for their use in the assembly of components, finished products, or spare parts. Parts with a longer lead time to build must be purchased, manufactured or obtained from an ETN source. Without a reliable and accurate ETN supply forecast, working capital continues to be tied up in purchased orders. This situation also results in carrying costs, risk of obsolescence, and general inefficiencies in the overall supply chain process.

The following sections of the paper are organized as follows. Section II briefly discusses existing reserve supply chain systems. Section III describes business challenges that were faced during this research. Section IV gives an overview of preparing and preprocessing of historical data followed by Section V which presents detailed analysis performed on historical data to make it compatible for modeling. Section VI discusses some experiments with some preliminary analysis that lead to final system methodology, which is discussed in Section VII, supported by results that are recorded in Section VIII. Finally, the paper is concluded in Section IX.

## II. Literature Review

The critical review of several Analytics-driven reverse supply chain researches which are relevant to this work are discussed in this section.

The returns disposition problem with outsourcing in reverse supply chains [10] expresses the same viewpoint, in designing a successful reverse supply chain. It is important to dispose returns within a short time, so that the returns can be processed more quickly as well as more profit can be obtained from them. As a result, a quick returns disposition can not only increase service level, but also get more profit for the manufacturing.

Most of the literatures provide us heuristics thought on fast reverse supply chain. Some of them are focused on product return flow to classify products before they returned to centralized evaluation center as well as how to optimize return logistics to save cost. A fast reverse supply chain can reduce the erosion of the value of return product; the quick returns disposition can get more profit for the manufacturing.

In optimizing the supply chain in reverse logistics [4] developed a components requirements planning procedure for end of life products, which used to find the most economical combination of products to disassemble in each period of the planning horizon. This procedure plays a guide role for our subject of study to realize analytics-based demand supply optimization.

Hampepure et. al [7] perform lifetime analysis for failure rate, the historical work-order data on assets be used; this information includes for each work order the time, location, asset number, work status and causes. The combined information is used as input in statistical modeling for failure rates and survival probability on each asset conditional on its current conditions and previous maintenance work orders. This Analytics methodology inspires us to bring more relevant historical data like machine sales and parts receipt in our return prediction instead of using return-alone data.

Methodology to forecast product returns for the consumer electronics industry [2] classifies returns in different reason code. There have been different distributions and return behavior for various reason code and forecast returns separately. These methods provide us an idea of grouping part numbers for forecast, the part numbers with same characters and specification be grouped into category and forecast the category instead of each part number. In our research, we also study the difference of returns behavior in different phase of return life-cycle and then use different model to forecast returns in different phase.

A lot of papers have been published related to speeding up reverse supply chain, but most of them focus on material flow of return product or component. Only a few of them focus on return information flow. Some papers forecast the failure or returns from field in life-cycle of product but the purpose is to prepare appropriate resource for the supply of replacement or maintenance. To differentiate with these research works, the key focus of our study is information flow, and predicting returns method be introduced into end of product life-cycle.

## III. Business Challenges

Conventional business models view parts return and reverse supply chain as a mandatory overhead cost of doing business. Companies typically operate their reverse supply chain by organizing forward supply chain manufacturing operations. This can cause misalignment due to several process problems, for example:

1) Lack of one single end-to-end process and a focus to drive improvements
2) Poorly developed or misaligned metrics
3) No end-to-end reverse supply chain data and tools to support decision management

The original materials returns process ensured the proper handling of returns at a consolidation center. A return occurs for various reasons including machine upgrades, warranty returns, end-of-lease and reseller returns. The original process had various pitfalls, e.g., the process owners or the procurement materials planning team could not predict when materials might be returned. This lack of information about returns inhibited the forecasting of future returns due to a high degree of uncertainty associated with it, thereby directly affecting the parts planning process.

Because of the lack of future returns forecasts, the ETN team was forced to wait until the consolidation centers had the physical equipment to know which parts were available for reutilization. Not knowing when ETN parts might be available caused churn in parts orders, affecting both manufacturing process and suppliers. Inventory levels of parts were also affected because parts were held in inventory that were not needed or could have been used elsewhere. This problem was magnified when it came to the recovery of high dollar value or high demand products and parts that were returned.

The original process was inefficient at handling scrap reclamation for parts that could not be recovered. To comply with existing and potential new regulations, an upgraded reclamation process was needed. As part of this system, new cost-efficient scrap reclamation processes were developed and deployed at manufacturing sites. These new processes ensure that environmental regulations are being met and material cost recovery is achieved. Business analytics provides insight into this decision making process by eliminating guesswork. The analytics solution creates a systematic way to balance economic and sustainability goals.

## IV. HISTORICAL DATA

The domain relevant historical data is gathered from various sources which consist of several important features cumulated on monthly basis including the number of machines shipped, number of upgrade requests, total parts returned etc. This section discusses operations that are performed on historical data before passing it for data modeling i.e., data preparation and pre-processing.

### A. Data Preparations

In data preparation phase, some business rules are applied on the raw data-source to make it compatible with the business requirements. Based on the rules, gross returns is considered as the target (dependent) variable and independent variables set includes machine shipments, upgrades, and new receipts. The rules are given below:

1) The instances of returns historical data that occur after general availability (GA) of next generation (N+1) are considered in the forecasting model
2) New receipts of 6 months prior to GA of N+1 generation excluded from the analysis and modeling, etc.

Certain transformations are applied on dataset features, which are given as follows:

1) 3, 6, etc. months of moving average on gross returns
2) Introduction of lagged period on shipments, upgrades and new receipts based on the GA of N+1 generation, because returns of generation N is usually triggered by GA of N+1

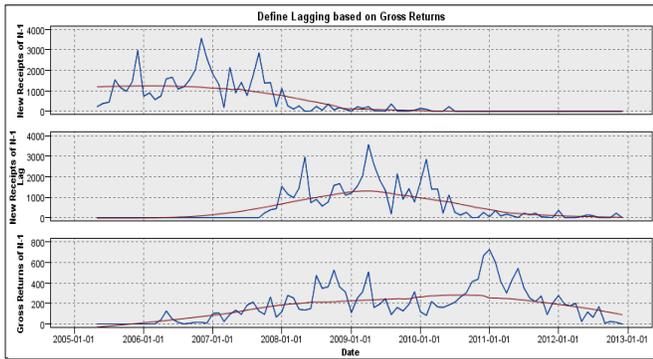

Figure 1. Lag period of 30 months is defined on new receipts to synchronize it with actual gross returns.

3) In some cases the historical data features do not match with the generation that is to be extrapolated. In order to make them compatible, a one-to-one mapping between features is applied on both datasets

The first two graphs of Figure 1 show trends of actual and lagged new receipts respectively while actual returns can be seen in last graph. A lag period of 30 months on new receipts is introduced because it is the GA of N+1 as well as instantiation of generation N returns.

### B. Data Preprocessing

The data pre-processing phase includes data cleansing by detecting and fixing anomalies in the dataset, magnitude normalization, and smoothing. This is considered as an important process of the system which prevents the overall system to generate over/under forecasted results.

*Outlier Detection*

Outliers or anomalies usually present wrong insights to the user. To prevent data from skewed analyses, anomalies are removed using specialized outlier detection algorithms, which search for unusual cases based upon deviations from similar cases and provide reasons for such deviations as well. Also, outliers vary by industry and by each case. In our case, some anomalies were found in the data due to new generation launch, seasonal promotion, financial crisis and its disruption on buying, returns behavior, etc. In most of the cases those anomalies contained solid business reasoning, so it was impossible to eliminate them; rather they are fixed by applying smoothing technique on the data.

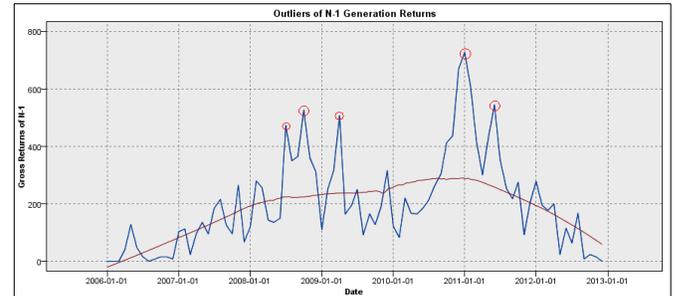

Figure 2. Five outliers of N-1 generation returns are encircled.

Equation (1) is used to extract outliers from each feature of the dataset separately.

$$Outlier_A = \begin{cases} Mean(feature_A) - 3 \times SD(feature_A) \leq feature\ A_i |_{i=1}^{n} \\ Mean(feature_A) + 3 \times SD(feature_A) > feature\ A_i |_{i=1}^{n} \end{cases} \quad (1)$$

*Normalization*

During the analysis of returns life-cycle, it is observed in most of the cases that the overall quantities are 3-4 times higher than the previous generation. The mean of generation N is found to be about 3-4 times greater than the previous generation. On the contrary, its standard deviation was 2-3 times less than N-1 generation. This problem of the mean being higher but the standard deviation being low was posing a hurdle to provide an accurate extrapolation of the current generation N. In order to deal with this, the difference is minimized and scaled the overall magnitude of historical data with the generation which has to be forecasted so that after model building it can be able to provide accurate forecasted numbers. Equation (2) is used to normalize each feature of generation N-1 with reference to N.

$$Normalization_A = \frac{\sum Generation\ N_A}{\sum Generation\ N-1_A} \times feature\ A_i\big|_{i=1}^n \quad (2)$$

The first graph in Figure 3 is showing the gross returns of generation N while the second and third graphs show actual N-1 and normalized N-1.

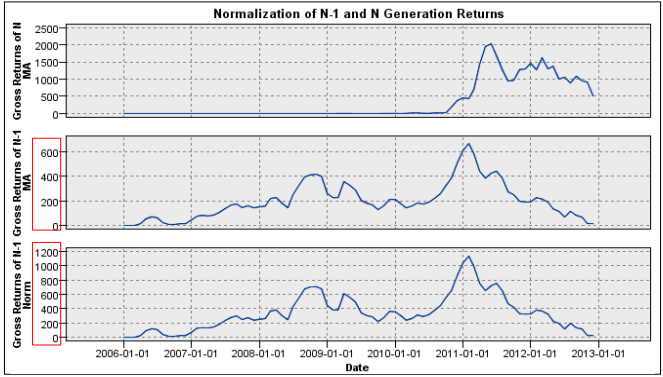

Figure 3. Magnitude normalization of N-1 generation with reference to the generation N.

*Smoothing*

Moving average smoothing technique is applied on the entire dataset to minimize the effect of spikes that is generated by very low and high numbers in the datasets. An average of the most recent 3, 6, etc. data points is computed. This process reduces the effect of spikes as well as makes the dataset features statistically significant for predictive modeling. Equation (3) is used to smooth each feature of dataset by applying 3 months moving average.

$$Moving\ Average_A = \times Average\left(feature_A\ instance_i\big|_{i=j}^{j+2}\right)\bigg|_{j=1}^n \quad (3)$$

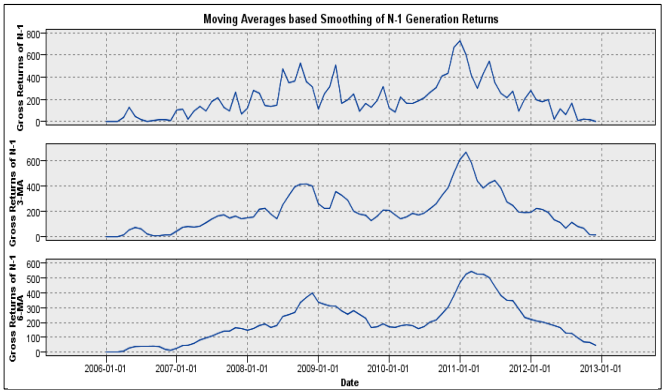

Figure 4. Moving averages based smoothing of N-1 generation gross returns.

Figure 4 shows moving averages based on smoothing of 3 and 6 months in the second and third graphs, while a lot of spikes can be observed in the first plot because it is the plot of actual returns without applying any smoothing.

## V. DATA ANALYSIS

This section presents the product returns life-cycle, correlation analysis, seasonality decomposition and causal factors of the features.

### A. Returns Life-cycle

The returns curve can be divided into three phases i.e., ramp-up, plateau/peak, and ramp-down. Each phase represents well-defined stages of the returns life-cycle as well as can be represented by calculating three separate equations of the line that can be used as a basis for predicting returns. Figure 5 illustrates the different sections of the life-cycle spreaded over the span of 32 months. The returns are usually triggered by general availability of N+1 generation and gradually attain the peak in around 15 months. Then partial non-linear peak phase usually remains there for next 10 months, followed by decline that ends up in the next 8-10 months.

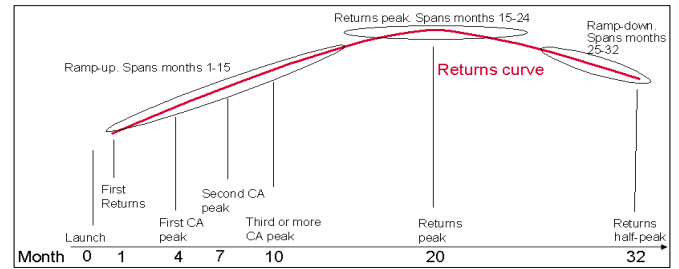

Figure 5. Three phase product returns life-cycle.

### B. Genealogy Study

Genealogy Study is conducted to find out historical data that is statistically most suitable for the forecasting of current generation. For that purpose, actual sales and gross returns of all previous generations are correlated with the same features of current generation to establish that the series are analogous. A previous generation is selected based on its high correlation with the current generation's data available by the time of modeling. This process is performed to get similar behavior over the span of entire product life-cycle. Figure 6 shows the analysis between current and previous generations for genealogy study.

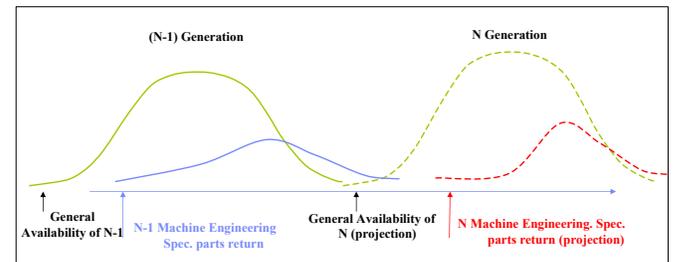

Figure 6. Analysis between current (N) and previous (N-1) generations.

## C. Correlation Analysis

Correlation analysis is performed between dependent and independent variables to pick the set of predictors which are strongly correlated with the target variable gross returns. The input feature set contains raw as well as several transformed features, e.g., lagged, moving averages, cumulative sum, etc. The correlation results between gross returns and transformed predictors are shown in Table I, where it can be observed that three months moving average of gross returns has better correlation with predictors then the raw gross returns. Furthermore, the transformation of shipments, upgrades and new receipts which are strongly correlated with the target are highlighted as bold font.

TABLE I
CORRELATION ANALYSIS BETWEEN DEPENDENT VARIABLE AND TRANSFORMED PREDICTORS

| Predictors | Gross Returns | | Gross Returns 3-MA | |
| --- | --- | --- | --- | --- |
| | Correlation | Strength | Correlation | Strength |
| Shipments | -0.134 | Weak | -0.118 | Weak |
| Shipments Lag 24 | 0.175 | Medium | 0.23 | Strong |
| Shipments Lag 30 | 0.208 | Strong | 0.28 | Strong |
| **Shipments Lag 42** | **0.33** | **Strong** | **0.357** | **Strong** |
| Shipments Lag 48 | 0.31 | Strong | 0.351 | Strong |
| Shipments Lag 54 | 0.313 | Strong | 0.339 | Strong |
| Upgrades | -0.139 | Weak | -0.137 | Weak |
| Upgrades Lag 24 | 0.188 | Strong | 0.245 | Strong |
| Upgrades Lag 30 | 0.184 | Medium | 0.274 | Strong |
| **Upgrades Lag 42** | **0.264** | **Strong** | **0.392** | **Strong** |
| Upgrades Lag 48 | 0.32 | Strong | 0.36 | Strong |
| Upgrades Lag 54 | 0.251 | Strong | 0.39 | Strong |
| New Receipts | -0.243 | Strong | -0.296 | Strong |
| **New Receipts CS** | **0.365** | **Strong** | **0.432** | **Strong** |

## D. Seasonality Decomposition and Causal Factors

Seasonality is suitable for data exhibiting a seasonal pattern as well as trends. It identifies the seasonal components in the past data and uses them in a forecasting model. In this case seasonal decomposition is applied on historical data generation as well as current generation which has to be extrapolated. Seasonality patterns are externally factored in features like shipments, upgrades, new receipts and gross returns.

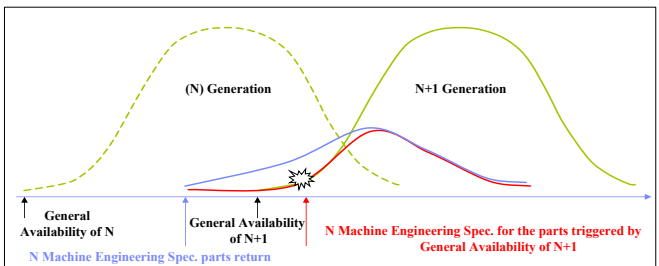

Figure 7. N-1 generation upgrades for the parts triggered by the general availability of generation N.

Following are some causal factors assessed as well as incorporated into the predicted model:
1) General availability of N+1 generation, which triggered ramp-up phase of generation N as shown in Figure 5.
2) General availability of N+2 generation, which triggered ramp-down phase of generation N.
3) Fire sale
4) Economic conditions
5) Engineering change (EC)
6) Some components may live cross-generations

## VI. EXPERIMENTATION

This section briefly discusses experiments that have been performed during the development of this system. There is a phase-wise modeling experiment performed to enhance the overall forecasted returns accuracy.

### A. Phase-wise Modeling

The phase-wise modeling was applied on the data to improve the overall accuracy of the system. In this experiment, the entire product returns life-cycle was split into three phases; i) Ramp-up (Phase-I), ii) Plateau/peak (Phase-II), and iii) Ramp-down (Phase-III) as shown in Figure 5. These three phases came-up from returns life-cycle analysis discussed in Section V (A). This approach is modeled using three separate models rather than one model for entire non-linear bell-shaped life-cycle.

*Phase-I: Ramp-up*

It is linear phase with positive slope and its duration is about 15 months. Linear regression algorithm was applied on this period of historical data because of its linear trend in the life-cycle.

*Phase-II: Plateau*

This phase is partially non-linear curve and usually remains for around next 8 to 10 months. Time series algorithm was applied to model this period of historical data.

*Phase-III: Ramp-down*

It is also linear phase but with negative slope and initiated from month 25 till GA of generation N+2. Again linear regression algorithm was applied on this period of historical data.

Each phase is triggered by a causal factor which is GA of N+1 and N+2 generations. The following methodology is applied to find out intersections of returns life-cycle:
1) The first interaction between phase-I and II is triggered by GA of generation N+1.
2) Similarly phase-II and III interaction is triggered by GA of N+2 generation.

## VII. METHODOLOGY

This section discusses overall methodology to build

ETN forecasting system, including predictive modeling, Early Warning Analytics (EWA), rule-based forecasting adjustments, model updating criteria and implementation.

*A. Predictive Modeling*

In data modeling phase, several supervised learning algorithms are used to forecast gross returns. Those learning algorithms are trained on entire product life-cycle and then extrapolated to generate forecasted returns of parts from a newer finished product.

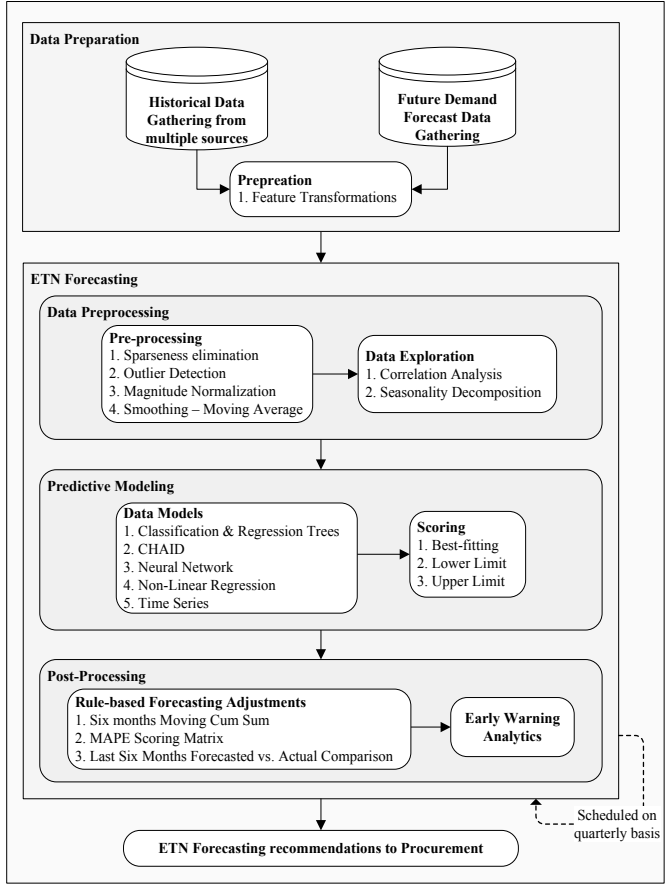

Figure 8. Detailed architecture of ETN forecasting system.

Figure 8 explains the detailed architecture of the system, which is further divided into several sections i.e., data preparation, preprocessing and exploration, modeling, 3-steps post-processing and preparation of results for decision making. The historical data is gathered from various sources and then feature transformation is applied on it to add more statistical significant features in the dataset. The updated dataset is passed to preprocessing module which applies various statistical operations to improve its quality including sparseness elimination, outlier detection, magnitude normalization and moving average based smoothing. This leads to more advanced statistical analysis on the updated dataset including correlation analysis and seasonal decomposition, to pick the most appropriate historical data generation as well as features which are strongly correlated to target variable - the gross returns.

Eventually clean dataset is forwarded to predictive analytics phase where several supervised learning algorithms are applied on it like C&RT, CHAID, Neural Network, non-linear regression and time series. These algorithms are evaluated and then ranked by higher correlation between predicted vs. actual returns as well as lower mean absolute percentage error (MAPE). The lower control intervals (LCI) and upper control intervals (UCI) are also computed with the best-fitted forecasted returns. These intervals become useful when best-fitted numbers generate over/under-forecasted returns.

A 3-step post-processing is applied on the output of modeling phase to benchmark its reliability. This process is named as Early Warning Analytics (EWA).

*B. Early Warning Analytics (EWA)*

It takes forecasted returns as input to generate a warning alert in case the scores are deviating from the actuals, enough to cross the defined threshold limit. The entire system runs automatically in batch mode, therefore, in case of over-forecasting, it becomes difficult to find out whether these deviations are due to discrepancies in predictive model or data-source quality. EWA overcome this issue by generating alerts on insignificant forecasts.

*Input Data*

The input data feed in EWA consists of four features including:
1) Previous cycle forecasted scores.
2) Current cycle forecasted scores.
3) The forecasted numbers selected by procurement planning team in previous cycle for decision making.
4) Actual gross returns till previous cycle.

*Methodology of EWA*

There is a 3-step methodology adopted to build EWA:
1) Calculate the deviation of the forecasted returns against the actuals over last three months using (4).

$$Deviation = Actual\ Returns - Previous\ cycle\ forecasted\ returns \quad (4)$$

If the deviations are negative then over-forecasting warning will be issued. This value is also used in (5) to calculate percentage absolute deviation (PAD) for recent three months.

$$Percentage\ Absolute\ Deviation = \frac{Deviation}{Actual\ Returns} \times 100 \quad (5)$$

2) Calculate the deviation of the forecasted returns again, but this time using previous cycle forecasted returns, which are used in decision making, against actuals over last three months using (4).

3) Take actions based on the outcome of the above mentioned calculations. In case of over-forecasting for the last 3-months cumulative sum, consider LCI returns or revisit the model with refreshed dataset. On the contrary, in case of under-forecasting of more than 20%, use UCI or revisit the model with updated dataset.

*C. Rule-Based Forecasting Adjustments*

Some rules are applied to adjust forecasting results generated by the supervised learning algorithms. The rules are:
1) Comparison of forecasted gross returns with actuals in the past 3 months prior to forecast decision point.
2) Adjustment of forecasted returns to minimize deviation beyond 10% over the last three months.
3) Adjustment of seasonality with 20% factor if the generation is still active to minimize the effect of spike.
4) Adjustment based on the products life-cycle, e.g., start of gross returns is considered from 12 months after GA to 24 months.

*D. Models Updating Criteria*

The model updating is scheduled on monthly basis because input dataset is refreshed on every month. Regardless of model scheduling frequency, the forecasting window is configured on quarterly basis but monthly results are used by EWA to check whether the forecasted returns are within the defined threshold.

*E. Implementation*

The implementation of this methodology is done in SPSS - a statistics computing language. The historical input dataset is feed into the system as flat comma separated files which are merged to create a single source and pass it to transformation process. It creates transformed features and then normalizes the magnitude of historical data with respect to the generation which has to be extrapolated.

The normalized data is then splitted into training (70%) and testing set (remaining 30%) and passed for predictive modeling which contains several supervised learning algorithms. These algorithms run on the preprocessed data simultaneously and rank based on MAPE and correlation between predicted vs. actual gross returns. The modeling process generates three predicted variables best-fitted, LCI and UCI of the algorithm which is ranked as the best. These three numbers are then used to calculate the MAPE.

Forecasted returns and their MAPEs are passed to 3-step process to validate their significance as well as raise an alarm in case of over-forecasting. After computing all the information, an output file is prepared to store the information in easily understandable format. The information contains actual gross returns, best-fitted forecasting, LCI, UCI, MAPE, and three step process numbers.

## VIII. RESULTS

The comprehensive benchmarking of the overall methodology is performed by using different evaluation methods. These methods include various statistics applied on the forecasted numbers to prevent reporting over-forecasted numbers, which was considered as major concern of the procurement planning team. Therefore the results are calculated in two different phases:
1) MAPE and correlation between actual vs. forecasted returns to select the best algorithm.
2) Early Warning Analytics (EAW) to check the validity of the forecast, which is explained in Section VII (C).

The MAPE of all the algorithms on the same dataset is computed in Table II including correlation between actual and forecasted numbers, LCI and UCI of forecasted numbers. Equation (6) is used to calculate the overall error of the system.

$$MAPE = Average(\frac{Actual - Forecasted}{Actual} \times 100) \quad (6)$$

From Table II it can be observed that CHAID outperformed the remaining algorithms, even though its correlation is ranked second; but overall this became the first choice that should be presented to the procurement planning team.

TABLE II
COMPARISON OF FIVE SUPERVISED LEARNING ALGORITHMS

| Algorithms | MAPE | | | Correlation |
| --- | --- | --- | --- | --- |
| | Best-fitted | LCI | UCI | |
| Linear Regression | 6.74 | 8.80 | 4.68 | 67.3 |
| C&R Tree | 8.96 | 14.92 | 3.00 | **99.00** |
| **CHAID** | **1.78** | **3.82** | **0.38** | 97.30 |
| Neural Network | 8.20 | 13.41 | 2.99 | 84.70 |
| Time Series | 2.51 | 25.36 | 30.39 | 95.80 |

After evaluating the system based on MAPE and correlation, EWA system is applied on the same best-fitted numbers to prevent reporting deviated forecasted numbers from the actuals. It calculates three different accuracies, for all five algorithms mentioned in Table II. Statistics which are part of EWA includes;
1) Average and standard deviation of six months basis moving cumulative sum of MAPEs,
2) Recent six months MAPE scoring matrix, and
3) Deviation of recent past six months actual and next six months forecasted returns.

Table II reports the comparison of five algorithms that is the outcome of EWA. Equations (7) and (8) are used to calculate MAPE scoring matrix, in which (7) presents the rules of assigning colors while (8) is used to output overall score of the algorithm. Equation (9) is the projection of future 6-12 months forecasted returns compared with recent past 6-12 months actual returns.

$$Red = < -10 \mid Yellow = Between\ 0\ to -10 \mid Green = > 0 \quad (7)$$

$$MAPE\ Scoring = 3 \times No.of\ Reds + 6 \times No.of\ Yellows + 3 \times No.of\ Green \quad (8)$$

Table III shows the evaluation of five algorithms based on the 3-step post-processing where the same algorithm, CHAID, performed significantly better than other candidate algorithms. Eventually this algorithm is deployed in production environment from where procurement team gathers updated forecasted returns on monthly basis which helps them to plan for next 12 months. This process significantly reduced supplier liability from 9 weeks to 12 months planning cycle which is quantified as 5% of 10 million US dollars liability.

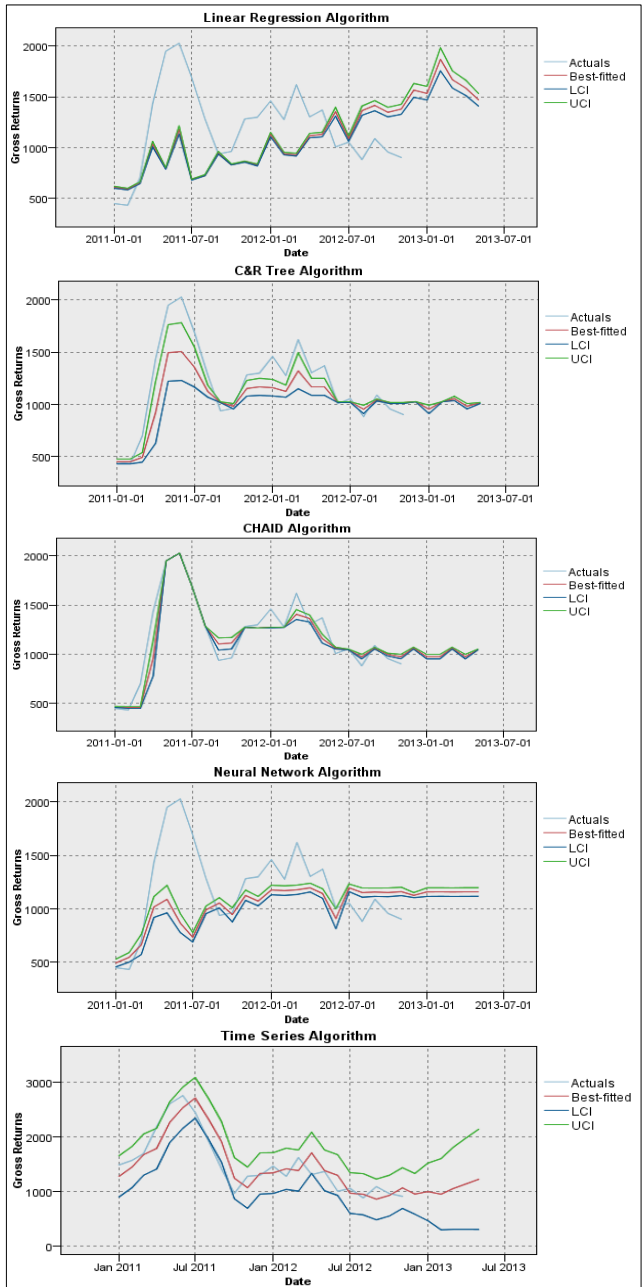

Figure 9. Magnitude Normalization of N-1 generation with reference to the generation N.

$$Actual\ vs\ Forecasted\ Projection = \sum(recent\ 6\ months\ Actuals) - \sum(next\ 6\ months\ Forecasts) \quad (9)$$

TABLE III
EVALUATION OF ALGORITHMS BASED ON 3-STEP POST-PROCESSING

| Algorithm | 6 Months Moving Cum Sum | | 6 Months MAPE Scoring | Actual vs Forecasted Projection |
|---|---|---|---|---|
| | Average | Std. Dev. | | |
| Linear Regression | 20.73 | 22.75 | -9 | 633.20 |
| C&R Tree | 12.41 | 7.48 | 15 | **27.31** |
| **CHAID** | **2.34** | **4.55** | **21** | 33.89 |
| Neural Network | 17.84 | 16.07 | -3 | 170.04 |
| Time Series | 3.34 | 6.14 | 9 | 67.75 |

IX. DISCUSSION AND CONCLUSION

Increased globalization, greater complexities of supply chains, and vast proliferation of data, have combined to accelerate the demand for business analytics. To compete globally, supply chain organizations need to adopt new smart analytics tools and processes to interpret readily available data, transform their business, drive improved decision making, and increase profits.

The ability to forecast returns prevents procurement from incurring inventory liability when ordering new parts. Procurement purchases parts before they are needed for their use in the assembly of components, finished products, or spare parts. Parts with a longer lead time to build must be ordered much further in advance of their need. Inventory liability increases when parts can be purchased, manufactured or obtained from an ETN source. Without a good visibility to ETN parts, fluctuations like unexpected inventory increases can occur because parts supply comes from ETN sources and new purchase orders.

This situation causes planning churn for related parts orders. Without a reliable and accurate ETN supply forecast, working capital continues to be tied up in purchase orders. This situation also results in carrying costs, risk of obsolescence, and general inefficiencies in the overall supply chain.

The forecasting solution incorporated SPSS to forecast the continuity of returns supply in a relatively short amount of time. Returns do not necessarily follow sales patterns or revenue projections that are commonly used in a forecasting process. Also, returns cannot be easily forecasted as salable items or marketing units can, by using more conventional demand forecasting techniques. By knowing what will be returning (in advance), a company can now reduce the expense of new parts purchases by using the forecasted returns as a trusted source of supply.

In addition to predicting the number of returns over time, the solution provides an upper and lower confidence interval of volume returns that are influenced by other business factors. A key benefit to this solution is the ability to predict a long-range returns pattern. From this result, the supplier can minimize supplier liability for high-cost parts with a longer manufacturing lead time. In addition, the supplier can create a new source of supply by accounting for the availability of forecasted returns.

The availability of forecasted returns enables procurement to take the following actions:

1) Authorize the recovery of the returning parts.
2) Release the parts to the recovery processes.
3) Reduce the total cost of bill of materials (BOM) that is needed to build new refurbished servers or service parts.

The solution is fully deployed as part of an end-to-end procurement business process with operational ownership. Operational business owners are responsible for executing and improving the solution over time and tracking its benefits. A set of metrics is updated after the execution of each planning cycle. These metrics enable the integrated supply chain management team to adjust the process and optimize the level of parts reutilization. Metrics are used in the following ways:

1) Assess the accuracy of the monthly returns forecast.
2) Measure the level of reused materials that are available for the Operational Commodity Manager to use.
3) Monitor the percentage of forecasted returns that are used by the Operational Commodity Manager.
4) Determine the risk level incurred in the total supply solution.

ACKNOWLEDGMENT

The author wishes to thank Raul Zeng, Jacob Thankamony, Jia Zhu and Harikrishna Aravapalli for taking part in system development, subject discussion and review of this work.

**Abbas R. Ali** is a Data Scientist working in IBM Business Analytics and Optimization Center of Competence. He received his B.S. degree in computer science and mathematics from the Institute of Management Sciences in 2004, M.S. degree in artificial intelligence and natural language processing from the National University of Computers and Emerging Sciences in 2009 and currently doing his PhD in artificial intelligence and predictive analytics from Bournemouth University. His current area of research is Meta-level Learning in the Context of Multi-component, Multi-level Evolving Predictive Systems.

Mr. Ali is a Chartered IT Professional from British Computer Society. He is author of several research papers in machine learning, statistics and predictive analytics areas. His current research interests include predictive and prescriptive, big data, and social network analytics.

**Dr. Pitipong J. Lin** is a Business Transformation Consultant for IBM Smarter Supply Chain Analytics. His expertise is in operations research, lean sigma, and supply chain strategy. He developed a patented solution for IBM Global Assets Recovery Services to enable timely economic-based reutilization decisions that help to expand global parts business and product refurbishment.

Dr. Lin has published over 30 journal articles and proceedings about reverse supply chain, analytics, and optimization. Dr. Lin received his M.S. degree in Management from the Boston University and a PhD. Degree in Industrial Engineering from Northwestern University in Massachusetts.